\documentclass[journal]{IEEEtranTIE}
\usepackage{graphicx}
\usepackage{cite}
\usepackage{picinpar}
\usepackage{amsmath}
\usepackage{url}
\usepackage{flushend}
\usepackage[latin1]{inputenc}
\usepackage{colortbl}
\usepackage{soul}
\usepackage{multirow}
\usepackage{pifont}
\usepackage{color}
\usepackage{alltt}
\usepackage[hidelinks]{hyperref}
\usepackage{enumerate}
\usepackage{siunitx}
\usepackage{breakurl}
\usepackage{epstopdf}
\usepackage{pbox}
\usepackage{xcolor}

\usepackage{amsmath,graphicx,siunitx}
\usepackage{float}
\usepackage{threeparttable}
\usepackage{booktabs}
\usepackage{makecell}
\usepackage{amssymb}
\usepackage{flushend}
\usepackage{hhline}
\usepackage{ulem}
\usepackage{threeparttable}
\newcommand\hy{\bgroup\markoverwith
	{\textcolor{yellow}{\rule[-.5ex]{2.1pt}{2ex}}}\ULon}
\newcommand{\cmark}{\ding{51}}
\newcommand{\xmark}{\ding{55}}

\begin{document}
\title{BAF-Detector: An Efficient CNN-Based Detector for Photovoltaic Cell Defect Detection}

\author{
	\vskip 1em
	
	Binyi Su, Haiyong Chen, and Zhong Zhou, \textit{Member, IEEE}
	
	\thanks{
		This work was supported by National Key R$\&$D Program of China under Grant 2018YFB2100601, National Natural Science Foundation of China under Grant 61872023, and National Natural Science Foundation of China under Grant 62073117. (\textit{Corresponding author: Zhong Zhou.}).
		
		B. Su, Z. Zhou are with the State Key Laboratory of Virtual Reality Technology and Systems, School of Computer Science and Engineering, Beihang University, Beijing 100191, China (e-mail: Subinyi@buaa.edu.cn; zz@buaa.edu.cn).
		
		H. Chen is with the School of Artificial Intelligence and Data Science, Hebei University of Technology, Tianjin 300130, China (e-mail: haiyong.chen@hebut.edu.cn).}
}

\maketitle
	
%\begin{abstract}
%The multi-scale defect detection for solar cell electroluminescence (EL) images is a challenging task, due to the feature vanishing as network deepens. To address this problem, a novel Bidirectional Attention Feature Pyramid Network (BAFPN) is designed by combining the novel multi-head cosine non-local attention module with top-down and bottom-up feature pyramid networks through bidirectional cross-scale connections, which can make all layers of the pyramid share similar semantic features. In multi-head cosine non-local attention module, cosine function is applied to compute the similarity matrix of the input features. Furthermore, a novel object detector is proposed, called BAF-Detector, which embeds BAFPN into Region Proposal Network (RPN) in Faster RCNN+FPN to improve the detection effect of multi-scale defects in solar cell EL images. Finally, some experimental results on a large-scale EL dataset including 3629 images, 2129 of which are defective, show that the proposed method performs much better than other methods in terms of multi-scale defects classification and detection results in raw  cell EL images.
%\end{abstract}

\begin{abstract}
	The multi-scale defect detection for photovoltaic (PV) cell electroluminescence (EL) images is a challenging task, due to the feature vanishing as network deepens. To address this problem, an attention-based top-down and bottom-up architecture is developed to accomplish multi-scale feature fusion. This architecture, called Bidirectional Attention Feature Pyramid Network (BAFPN), can make all layers of the pyramid share similar semantic features. In BAFPN, cosine similarity is employed to measure the importance of each pixel in the fused features. Furthermore, a novel object detector is proposed, called BAF-Detector, which embeds BAFPN into Region Proposal Network (RPN) in Faster RCNN+FPN. BAFPN improves the robustness of the network to scales, thus the proposed detector achieves a good performance in multi-scale defects detection task. Finally, the experimental results on a large-scale EL dataset including 3629 images, 2129 of which are defective, show that the proposed method achieves 98.70$\%$ (F-measure), 88.07$\%$ (mAP), and 73.29$\%$ (IoU) in terms of multi-scale defects classification and detection results in raw PV cell EL images.
\end{abstract}


\markboth{IEEE TRANSACTIONS ON INDUSTRIAL ELECTRONICS}%
{}

\definecolor{limegreen}{rgb}{0.2, 0.8, 0.2}
\definecolor{forestgreen}{rgb}{0.13, 0.55, 0.13}
\definecolor{greenhtml}{rgb}{0.0, 0.5, 0.0}

\section{Introduction}

\IEEEPARstart{T}{he} multicrystalline photovoltaic (PV) cell defects will lead to a seriously negative impact on the power generation efficiency. Moreover, these defective cells will generate a lot of heat in the process of power generation, which may lead to fire and property loss \cite{Alvaro2020}. Therefore, automated defect detection based on computer vision plays a vital role in the manufacturing process of PV cells. This process can advantage safe and high-efficiency operation of the large-scale PV farms. 

\begin{figure}[!t]
	\centering
	\includegraphics[width=7.5cm]{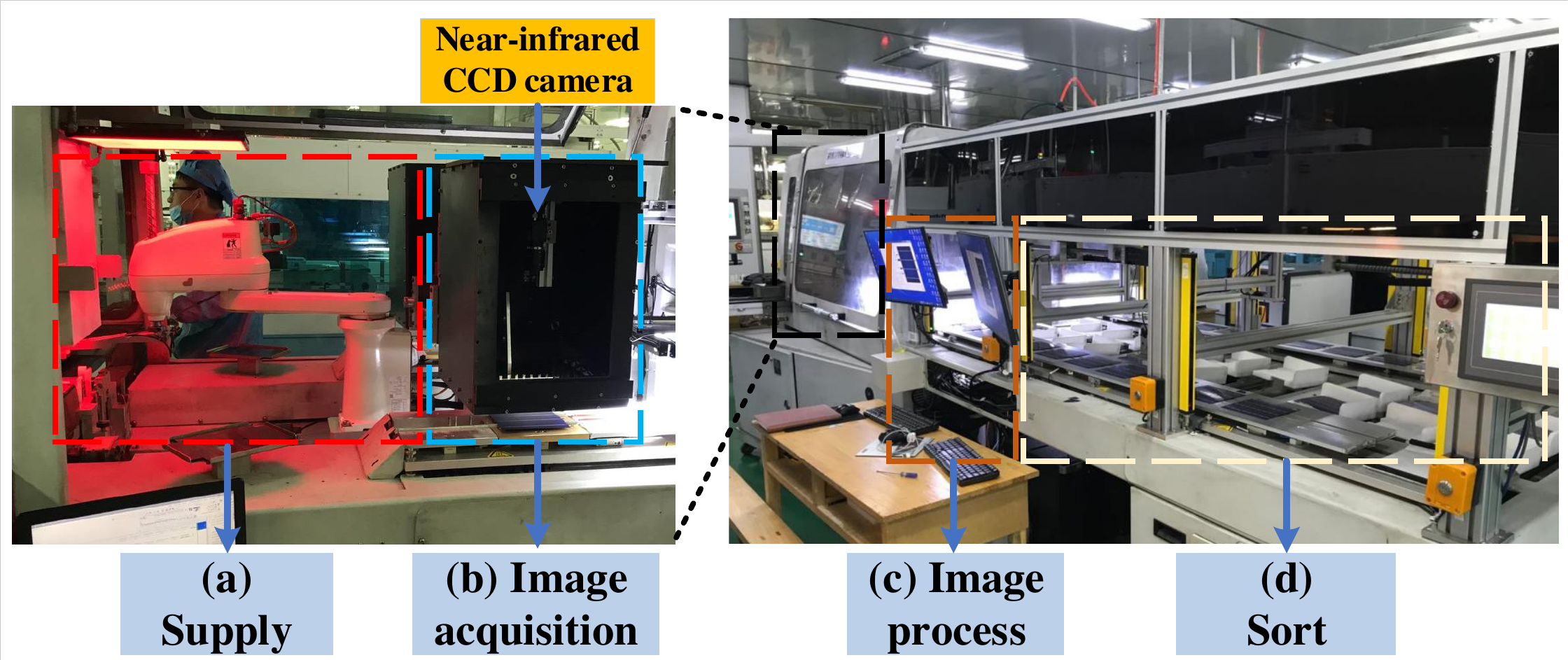}
	\caption{PV cell intelligent defect detection system.}\label{fig1}
\end{figure}

\begin{figure}[!t]
	\centering
	\includegraphics[width=7.7cm]{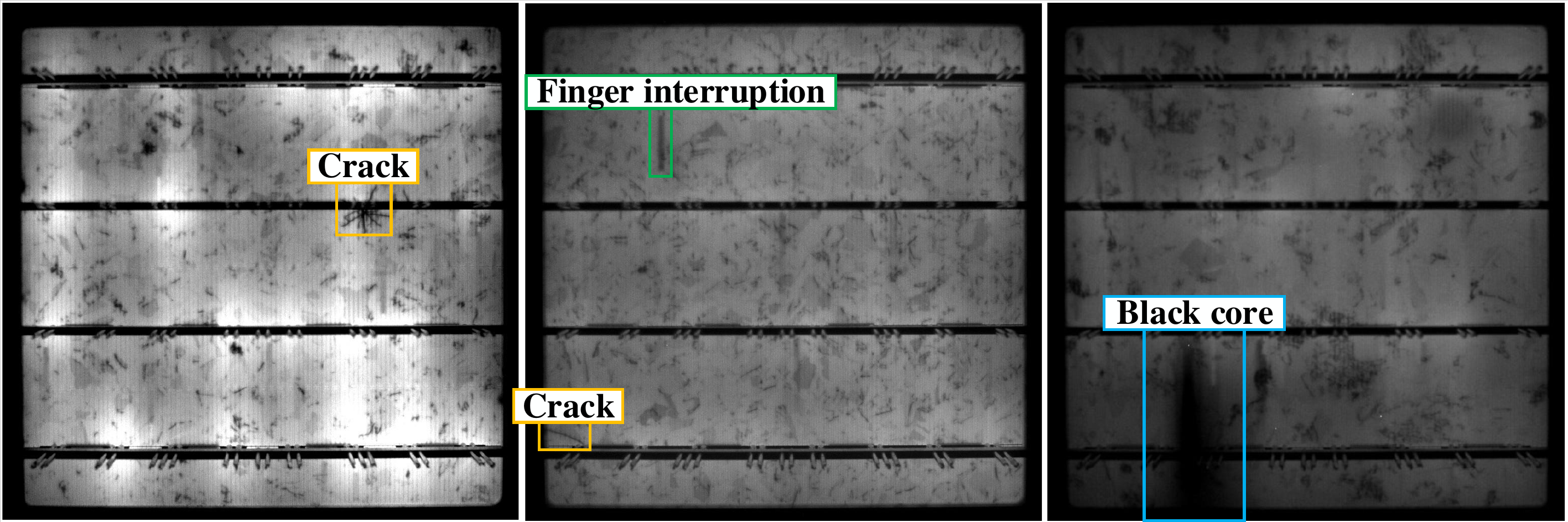}
	\caption{Three raw EL near-infrared images with two crack
		defects in yellow boxes, one finger interruption defect in green
		box, one black core defect in blue box.}\label{fig2}
\end{figure}

PV cell defect detection aims to predict the class and location of multi-scale defects in an electroluminescence (EL) near-infrared image 
\cite{2009Fuyuki, Mahmoud2019}. It is captured and processed by the following defect detection system, which integrates various sensors such as leakage circuit breaker to achieve safe and efficient fault elimination of PV cells. As is shown in Fig. \ref{fig1}, this intelligent defect detection system contains four components: supply subsystem, image acquisition subsystem, image process subsystem, and sort subsystem. By the image acquisition subsystem, the internal defects of PV cells that cannot be directly seen by the naked eye are clearly presented to us, as shown in Fig. \ref{fig2}. Regions of crystal silicon with higher conversion efficiency exhibit brighter luminescence in the sensed image. However, defects appear as dark regions because they are inactive and cannot emit light. Fig. \ref{fig2} presents three raw EL near-infrared images with three types of defects: crack, finger interruption and black core under complex background disturbance. The crack defect mainly includes line-like and star-like shapes, which show random texture distribution and multi-scale characteristic. The finger interruption defect presents strip-like shape and vertical distribution. Black core defect presents blob-like shape and is a cluster of black region in EL image. Except for the defects, the dislocation and four busbars also present as dark regions and may overlap with the defects sometimes. It causes automatic defect detection in PV cell EL images extremely difficult. 

To automatically identify these defects in EL image, many conventional computer vision-based methods \cite{2016Demant,2019Su} have been proposed to satisfy the urgent demand of the quality monitoring. Handcrafted features and data-based classifiers \cite{Firuzi2019,Luo2019} are usually the main tools, which use feature descriptors to generate the feature vectors of texture, color, shape, and spectral cues, and then, adopt classifiers to realize defects inspection. Feature extraction in conventional methods mainly relies on manually designed descriptors. This process requires professional knowledge and a complicated parameter adjustment. Moreover, each method targets to a specific application and has poor generalization ability and robustness.

\begin{figure}[!t]
	\centering
	\includegraphics[width=8.5cm]{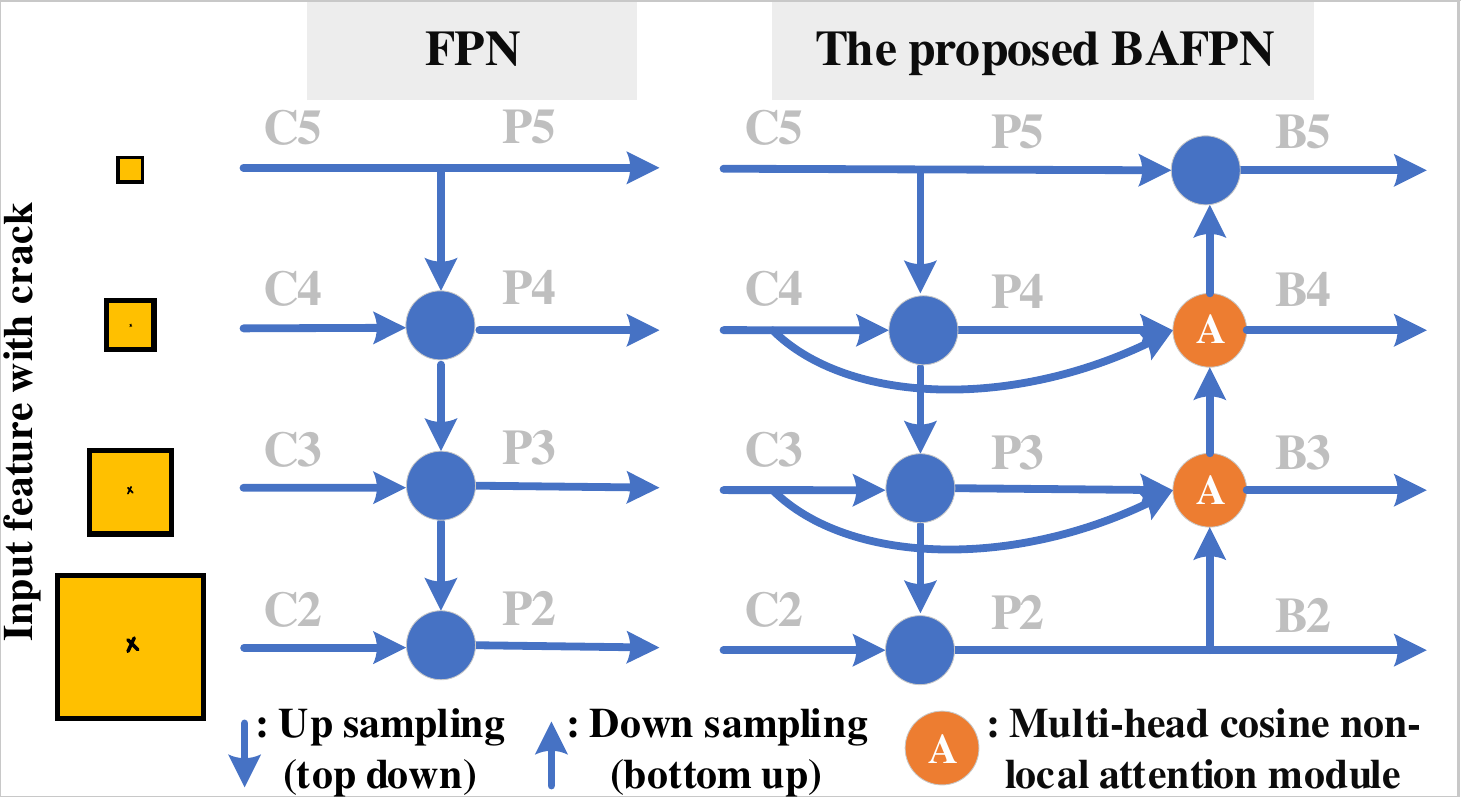}
	\caption{The architecture of Feature Pyramid Network and the proposed Bidirectional Attention Feature Pyramid Network (BAFPN).}\label{fig3}
\end{figure}

Recently, deep learning models, particularly the convolutional neural network (CNN), have received substantial interest in industrial applications \cite{2018NB,2020Hu,2019Sassi,2019Akram,Pang2020,2020Han,2020FFCNN,2020Su}. Deep learning is a data-driven method. It presents the advantages of high accuracy, wide versatility, and strong plasticity. According to the learning of a large number of samples, the specific feature representation of the dataset can be obtained. The deep learning models are more robust and have more generalization ability. However, with the deepening and downsampling of the network during the training process, the features of small defects such as micro-crack, finger interruption are very easy to vanish. Poor feature representation will lead to poor detection result. Thus, retaining the features of these small defects in the process of network deepening is crucial to improve the effectiveness of multi-scale defect detection task. Feature Pyramid Network (FPN) \cite{2017FPN} has the potential to solve this challenging problem. As shown in Fig. \ref{fig3}, the shallow layers contains much more small defect features than the deep layer, FPN integrates pyramid features of different scales from top to bottom, which is very beneficial for small defect detection. Subsequently, bottom-up path augmentation (PA-Net) \cite{2018PANet,2020efficientnet} is proposed to enhance the entire feature hierarchy with accurate localization signals in lower layers. This method  shortens the information path between lower layers and topmost feature.

Inspired by above excellent works \cite{2017FPN,2018PANet,2020efficientnet}, Bidirectional Attention Feature Pyramid Network (BAFPN) is proposed in this paper, which is a multi-scale network that can be applied to pyramidal feature fusion. Different from above two feature fusion modules (FPN and PA-Net), BAFPN innovatively adopts a novel multi-head cosine non-local attention module to capture the refined information of different scales, which is beneficial to the bottom-up feature transfer. Moreover, the residual connection from the original feature to the multi-head cosine non-local attention module can guarantee the fully exploit of the same scale features. Finally, a BAF-Detector is proposed, which embeds BAFPN into Region Proposal Network (RPN) in Faster RCNN+FPN \cite{2017FPN} to improve the detection effect of multi-scale defects in PV cell EL images. The main contributions of this paper are as follows.

\begin{enumerate}
	\item A multi-head cosine non-local attention module is proposed by employing cosine function to calculate the feature similarity, which can achieve the highlighting of defect features and the suppression of background features simultaneously. 
	\item An Bidirectional Attention Feature Pyramid Network (BAFPN) is proposed to boost the bottom-up feature transfer, which is formed by combining multi-head cosine non-local attention module with top-down and bottom-up feature pyramid networks through bidirectional cross-scale connections. BAFPN improves the robustness of the network to scales, which are beneficial to multi-scale defect detection.
	\item An object detector, called BAF-Detector, is proposed, which embeds BAFPN into Region Proposal Network (RPN) in Faster RCNN+FPN. BAF-Detector has significantly improved the detection effect of multi-scale defects in EL images, which can meet the high-quality requirements of PV cell industrial production.
	%	\item We propose a largest VOC2007 format photovoltaic solar cell EL image dataset, which contains 2000 defect-free raw EL images and 2127 defective EL images of crack, finger interruption, black core defects that have been manually labeled by ourselves with guidance of the experienced workers.
\end{enumerate}

This paper is organized as follows:  Section \uppercase\expandafter{\romannumeral2} presents an overview of the related works. Section \uppercase\expandafter{\romannumeral3} gives the details of the proposed methods. Section \uppercase\expandafter{\romannumeral4} presents extensive experiments and ablation studies. Finally, Section \uppercase\expandafter{\romannumeral5} concludes this article.

% Please add the following required packages to your document preamble:
% \usepackage[table,xcdraw]{xcolor}
% If you use beamer only pass "xcolor=table" option, i.e. \documentclass[xcolor=table]{beamer}
\begin{table*}[]
	\caption{The Comparison of the Proposed BAF-Detector with the Related Methods.}
	\centering
	\label{table_0}
	\begin{threeparttable}
	\begin{tabular}{cccccc}
		\hline\hline
		{\color[HTML]{31353B} Methods}                 & {\color[HTML]{31353B} Defect type}   & {\color[HTML]{31353B} CNN model}       & {\color[HTML]{31353B} Network scale} & {\color[HTML]{31353B} Classification}                                                                                                        & {\color[HTML]{31353B} Detection}                                                                         \\ \hline
		{\color[HTML]{31353B} Deitsch \textit{et al.} \cite{2019Deitsch}} & {\color[HTML]{31353B} Internal (EL)} & {\color[HTML]{31353B} VGG16}           & {\color[HTML]{31353B} Single}        & {\color[HTML]{31353B} Defect, defect-free}                                                                                                   & {\color[HTML]{31353B} No}                                                                                \\
		{\color[HTML]{31353B} Chen \textit{et al.} \cite{Chen2020}}    & {\color[HTML]{31353B} surface}       & {\color[HTML]{31353B} AlexNet}         & {\color[HTML]{31353B} Single}        & {\color[HTML]{31353B} Defect, defect-free}                                                                                                   & {\color[HTML]{31353B} No}                                                                                \\
		{\color[HTML]{31353B} Du \textit{et al.} \cite{2020Du}}      & {\color[HTML]{31353B} surface}       & {\color[HTML]{31353B} GoogleNet}       & {\color[HTML]{31353B} Single}        & {\color[HTML]{31353B} \begin{tabular}[c]{@{}c@{}}Broken edge, surface impurity, scratch,\\  hot spot, crack, large area damage\end{tabular}} & {\color[HTML]{31353B} No}                                                                                \\
		{\color[HTML]{31353B} Han \textit{et al.} \cite{2020Han}}     & {\color[HTML]{31353B} surface}       & {\color[HTML]{31353B} RPN+UNet}        & {\color[HTML]{31353B} Single}        & {\color[HTML]{31353B} Defect, defect-free}                                                                                                   & {\color[HTML]{31353B} defect}                                                                            \\
		{\color[HTML]{31353B} Su \textit{et al.} \cite{2020Su}}      & {\color[HTML]{31353B} Internal (EL)} & {\color[HTML]{31353B} Faster RPAN-CNN} & {\color[HTML]{31353B} Single}        & {\color[HTML]{31353B} \begin{tabular}[c]{@{}c@{}}Crack, finger interruption, \\ black core, defect-free\end{tabular}}                        & {\color[HTML]{31353B} \begin{tabular}[c]{@{}c@{}}Crack, finger interruption,\\  black core\end{tabular}} \\
		{\color[HTML]{31353B} This paper}              & {\color[HTML]{31353B} Internal (EL)} & {\color[HTML]{31353B} BAF-Detector}    & {\color[HTML]{31353B} Multi}         & {\color[HTML]{31353B} \begin{tabular}[c]{@{}c@{}}Crack, finger interruption, \\ black core, defect-free\end{tabular}}                        & {\color[HTML]{31353B} \begin{tabular}[c]{@{}c@{}}Crack, finger interruption,\\  black core\end{tabular}} \\ \hline\hline
	\end{tabular}
    \end{threeparttable}
\end{table*}

\section{RELATED WORKS}
\subsection{Automatic Defect Detection Methods}
Hand-crafted feature based algorithm usually extract the image features such as color, texture, shape, and spatial relationship to train and test the discriminant classifiers for defect recognition. Demant \textit{et al.} \cite{2016Demant} integrated four descriptors features to detect micro-crack defect in photoluminescence (PL) and EL images. This method greatly enhanced the feature representation ability to the defect. Su \textit{et al.} \cite{2019Su} introduced a novel gradient-based descriptor to extract the defect gradient feature in EL image. They also adopted similarity analysis and clustering to capture the image global information. This method achieved promising results on EL-2019 dataset. Despite these methods achieving promising performance in specific situations, most are heavily rely on the expertise and cannot transfer directly between different applications.  

At present, convolutional neural network (CNN) has become one of the most popular research direction in the field of computer vision \cite{2018NB,2020Hu,2019Sassi,2019Akram,2020Han,2020Su}. CNN model has achieved a great success in image classification \cite{2018NB,2019Akram}, detection \cite{2020Su,2020Zhang,Liu2020}, segmentation \cite{2019Nakazawa,2019Fu} and reconstruction \cite{cycle2020} tasks. Thus, applying CNN to detect PV cell defect in EL image is of good prospect. Deitsch \textit{et al.} \cite{2019Deitsch} presented a novel approach using light CNN architecture for recognizing defects in EL images, which required less computational power and time. Chen \textit{et al.} \cite{Chen2020} proposed a weakly supervised CNN model with attention mechanism to accomplish surface defect classification and segmentation in PV cell image.  Du \textit{et al.} \cite{2020Du} introduced an CNN-based algorithm for efficient and innovative defect detection in existing industrial production line. This approach achieved successful application in Si-PV cell defects classification and detection task. Han \textit{et al.} \cite{2020Han} proposed a defect segmentation method for polycrystalline silicon wafer based on deep learning. This method applied Region Proposal Network (RPN) to generate the underlying defect boxes, then a segmentation network was employed for pixel-wise defect region division. Su \textit{et al.}\cite{2020Su} proposed a novel Region Proposal Attention Network (RPAN) to detect defect in raw PV cell EL images. RPAN employed the attention mechanism to refine the CNN-extracted feature maps, which significantly enhance the  classification and detection performance. However, the detection effect of this method for multi-scale defects is not satisfactory, especially for micro-crack, which motivates this study. 

\begin{figure}[!t]
	\centering
	\includegraphics[width=6.26cm]{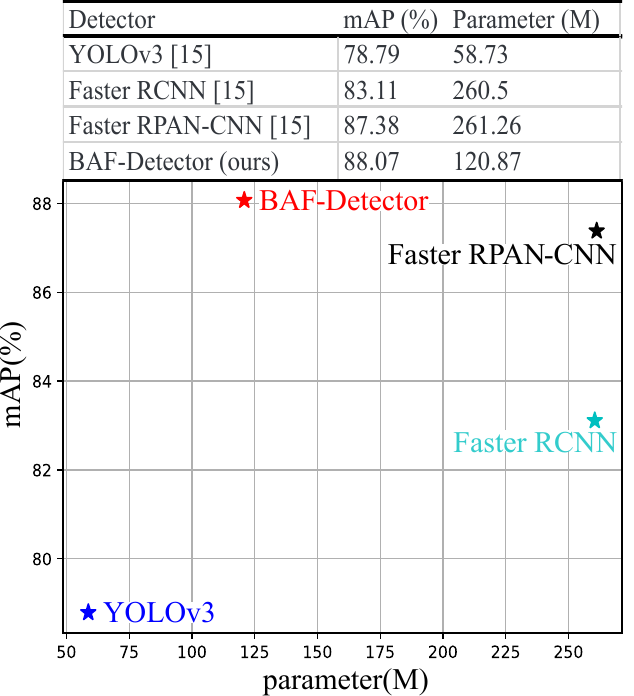}
	\caption{The comparison of BAF-Detector with Reference \cite{2020Su}.}\label{fig3-1}
\end{figure}

The comparison of the proposed BAF-Detector with the related methods are shown in Table \ref{table_0} and Fig. \ref{fig3-1}. As can be seen in Table \ref{table_0}, Reference \cite{2020Su} is closest to this paper. The difference between Reference \cite{2020Su} and this paper is that this paper is multi-scale network, which can improve the robustness of the network to scales. Furthermore, a quantitative demonstration and a detailed comparison with the experimental results of Reference \cite{2020Su} is shown in Fig. \ref{fig3-1}. The mean Average Precision (mAP) of BAF-Detector reaches 88.07$\%$ with fewer parameter number than Faster RCNN and Faster RPAN-CNN. Although the number of parameters is more than YOLOv3, the mAP value is 9.28$\%$ higher than it. More experimental results are shown in Table \ref{table_3}.

\subsection{CNN-based Detectors}
The CNN-based detectors can be roughly classified into the following two types: regression-based approaches, (e.g., Retinanet \cite{2020Retinanet}, YOLOv3\cite{2018YOLOv3}) and region-based approaches, (e.g., Faster RCNN \cite{2017Faster}, Mask RCNN \cite{2017Mask}). YOLOv3 is the most popular regression-based algorithm. It treats object detection as a regression problem, and then predicts the location and category of object. Regression-based approaches present fast detection speed, but the inspection accuracy is not satisfactory. As for region-based methods, many candidate proposals are firstly generated by Region Proposal Network (RPN), then these proposals are further classified and regressed by the following deep network. Region-based methods present high detection accuracy, but demand slightly more computational complexity. PV cell defect detection requires high detection accuracy to prevent defective cells from entering the next manufacturing process. Thus, to achieve a high-accuracy performance of defect detection in PV cell EL image, an region-based method is proposed to identify the fault cells in this paper. 

\subsection{Attention Mechanism}
The attention module judges the importance of each pixel in the input CNN features. The pixel importance of the target region is higher than that of the background region. Thus, the attention module can focus on learning the target area through large weigh assignment. As an embedded sub-network, attention module has been widely applied to various tasks \cite{2020Su,wang2018nonlocal,2019Shen,zhou2019ssacnn,Tang2021}. Wang \textit{et al.} \cite{wang2018nonlocal} proposed a non-local attention model to capture the contextual information of the object, which can boost feature representation ability of the image segmentation task. Shen \textit{et al.} \cite{2019Shen} employed channel-wise attention model to suppress the noise features of background and guide multi-scale feature fusion in object detection task. Zhou \textit{et al.} \cite{zhou2019ssacnn} applied visual attention model and wavelet transform to detect defects in glass bottle bottom. Tang \textit{et al.} \cite{Tang2021} presented a new spatial attention bilinear CNN to detect defect in X-ray images of castings. Su \textit{et al.} \cite{2020Su} proposed a novel Complementary Attention Network (CAN), which not only can suppress noise features of background, but also focuses on spatial locations of the defects in the PV cell EL images. In this paper, compared with the traditional non-local attention module, the multi-head cosine non-local attention module can better achieve the highlighting of defect features and the suppression of background features through cosine similarity calculation.

\section{METHODOLOGY}
In this section, we firstly introduce the proposed multi-head cosine non-local attention module, and then the architecture of the proposed BAF-Detector including BAFPN block is presented in details.

\begin{figure}[!t]
	\centering
	\includegraphics[width=7cm]{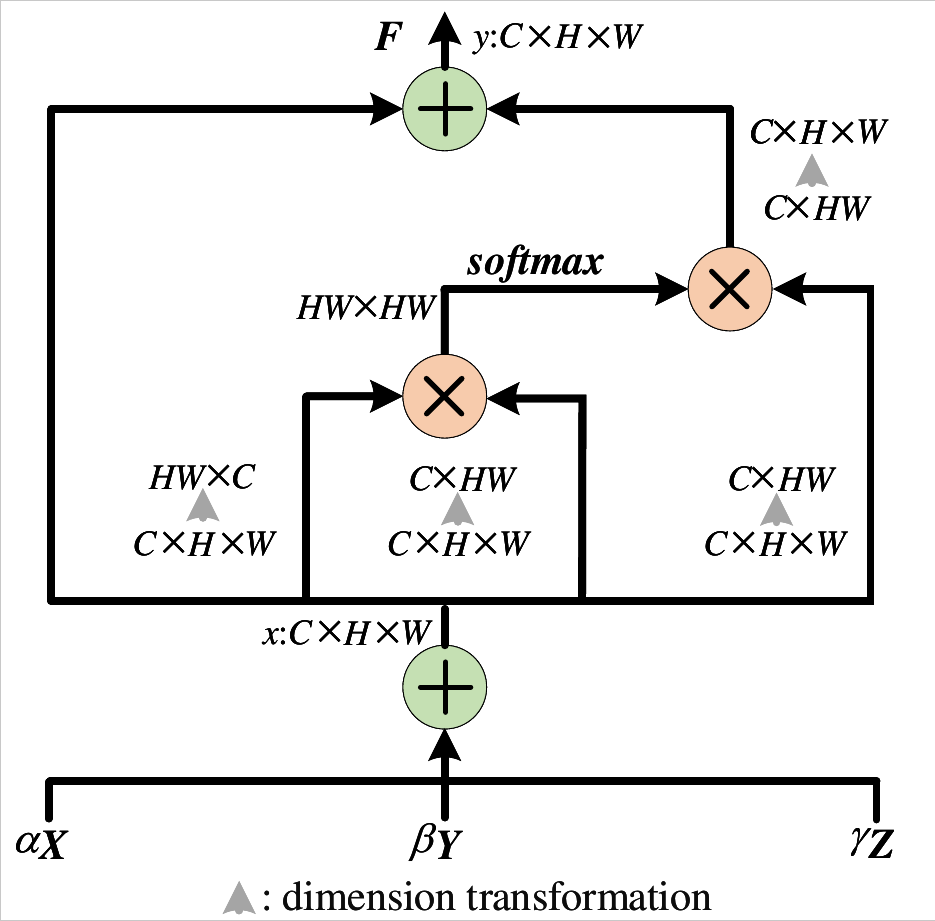}
	\caption{Architecture of the proposed multi-head cosine non-local attention module. $\oplus$ donates the element-wise addition, $\otimes$ donates the matrix multiplication.}\label{fig4}
\end{figure}

\subsection{The Proposed Multi-head Cosine Non-Local Attention Module}
The flowchart of multi-head cosine non-local attention module is illustrated in Fig. \ref{fig4}. Discriminant feature representations for defect and background in EL image are essential to fault identification. It can be accomplished by employing the attention module to highlight the defect feature and suppress the complicated background feature simultaneously. As for multi-head cosine non-local attention module, it depends on the cosine feature similarity matrix multiplying with the original input features to realize useful feature emphasize and disturbed feature suppression. 

As illustrated in Fig. \ref{fig4}, the multi-head cosine non-local attention module represents that three feature spaces $X\in\mathbb{R}^{C\times W\times H}$, $Y\in\mathbb{R}^{C\times W\times H}$, and $Z\in\mathbb{R}^{C\times W\times H}$ are input into the attention module ($C$, $W$, and $H$ are the channel number, width and hight of the input feature maps respectively). $\alpha$, $\beta$, and $\gamma$ are the balance factors that are employed to reweight with each head respectively, which are all equal to 0.5 in this paper. The fused input feature $x$ is defined as:

\begin{equation}\label{eq4}
	x=\alpha X+\beta Y+\gamma Z
\end{equation}

The feature $x\in\mathbb{R}^{C\times W\times H}$ will be converted to $\mathbb{R}^{C\times N}$ through dimension transformation, where $N=WH$ represents the total position number in previous feature map. After that, the transposition of $x$ will multiply with itself to calculate the cosine similarity map $f$. The similarity computing in original non-local attention module \cite{wang2018nonlocal} is defined as:

\begin{equation}\label{eq1}
	f(x_i,x_j)=x_i^Tx_j
\end{equation}
where $i$ and $j$ are the index of a position in the input feature map $x\in\mathbb{R}^{C\times W\times H}$. Here $x_i^Tx_j$ is dot-product similarity, which will be normalized by the following $softmax$ operation to form the attention map. $softmax$ used for element normalization represents the ratio of $exp$ index of the element to the sum of $exp$ indexes of all elements. It is performed on each row of the similarity map, and is defined as follows:
\begin{equation}\label{eq2}
	s_{j,i}={\textstyle\frac{exp(f(x_i,x_j))}{\sum_{i=1}^Nexp(f(x_i,x_j)))}}
\end{equation}
where $s_{j,i}$ measures the impact of $i^{th}$ position on $j^{th}$ position.   There is a question for the dot-product similarity computing of non-local attention module, once most of numerator $exp(f(x_i,x_j))$ are big values, the denominator $\sum_{i=1}^Nexp(f(x_i,x_j))$ will be very large compared with numerator. Then, $s_{j,i}$ is close to zero, the difference between each element in $s$ is very small. Thus, $s_{j,i}$ cannot be used to re-weight with the feature $g(x)$, where $g(x)$ is the output of a 1$\times$1 convolution operation. To solve this problem, dot-similarity is replaced by the cosine similarity, which will limit all elements $f(x_i,x_j)$ to [-1, 1]. After being normalized by $softmax$, these elements are suitable to keep their differences. The cosine similarity matrix is defined as: 

\begin{figure}[!t]
	\centering
	\includegraphics[width=8.7cm]{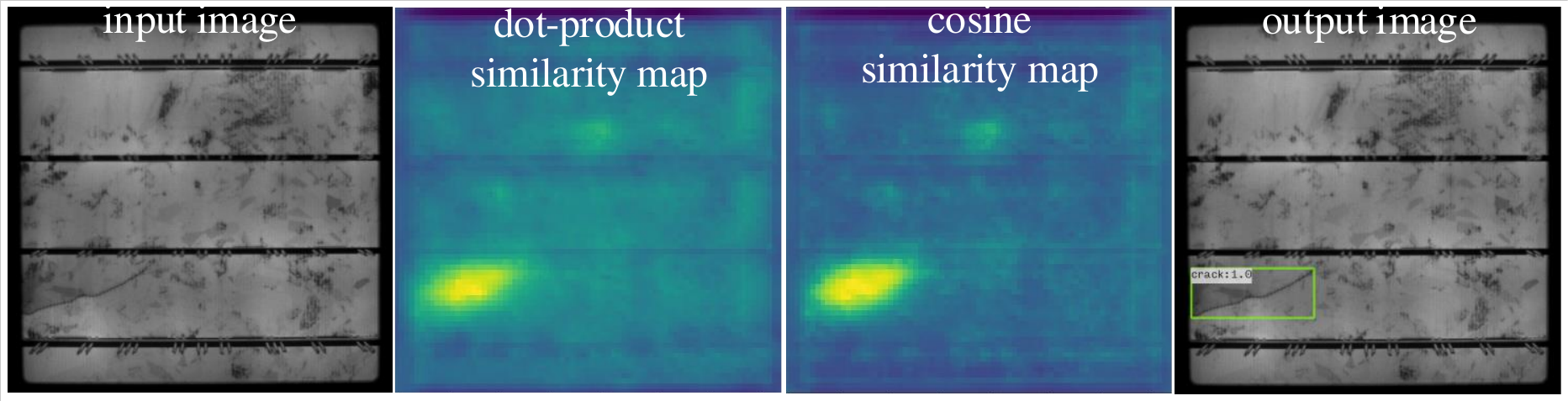}
	\caption{Visualization of dot-product similarity map and cosine similarity map in B3 layer of the proposed BAF-Detector.}\label{fig5}
\end{figure}

\begin{figure}[!t]
	\centering
	\includegraphics[width=8.7cm]{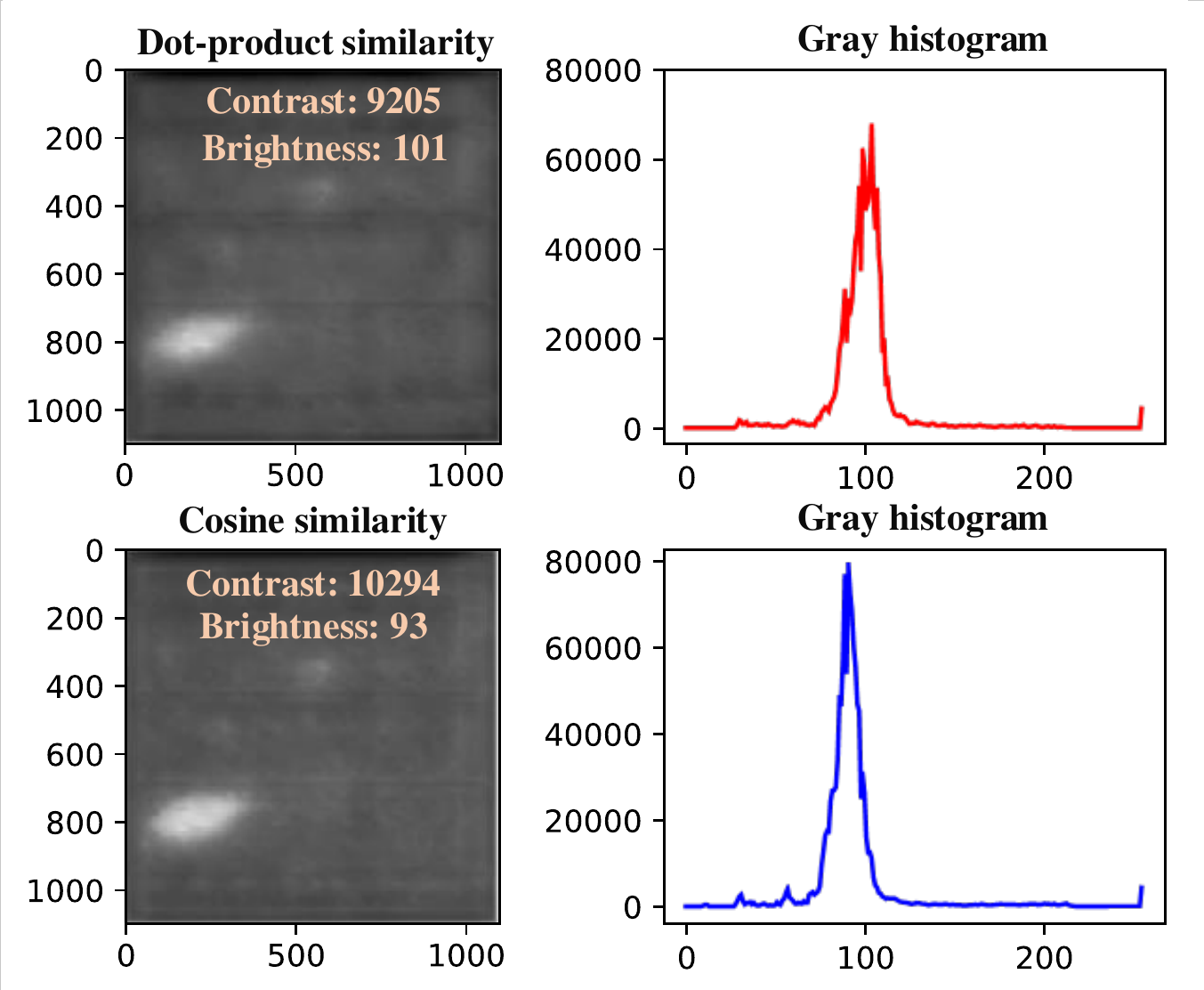}
	\caption{Contrast, brightness and gray histogram of the dot-product similarity map and cosine similarity map.}\label{fig5_1}
\end{figure}

\begin{equation}\label{eq3}
	f(x_i,x_j)=\frac{x_i^Tx_j}{\left\|x_i\right\|\left\|x_j\right\|}
\end{equation}
where $\left\|x_i\right\|$ and $\left\|x_j\right\|$ are the norms of vectors $x_i$ and $x_j$ respectively. Noting that cosine similarity can make non-local attention acquire more discriminant features, which achieves better performance than dot-similarity to highlight the useful feature and suppress the complex background feature, as shown in Fig. \ref{fig5}. It is not difficult to see that the cosine similarity map is better than the dot-product similarity map, whether the emphasize of defect feature or the suppression of complex background. Moreover, as shown in Fig. \ref{fig5_1}, contrast, brightness and gray histogram are applied to quantify the effect of the improvement. The high contrast and low brightness indicate that the defect features are highlighted and the complex background features are suppressed. The gray histogram of the cosine similarity map is narrower and higher than the dot product similarity map, which illustrates that the important area of the cosine similarity map is mainly concentrated on the defect.

Refer to Eq. \ref{eq2}, the $softmax$ function is employed to normalize the similarity map to form the attention map, which will multiply with the re-dimension feature $g(x)\in\mathbb{R}^{C\times N}$. Finally, the dimension of the result will be transformed to $\mathbb{R}^{C\times W\times H}$, and perform a element-wise sum operation with the input features $x$. The output $y\in\mathbb{R}^{C\times W\times H}$ of the multi-head cosine non-local attention module is defined as:

\begin{equation}\label{eq5}
	y_j=\sum_{i=1}^{N}s_{j,i}g(x_i)+x_i
\end{equation}

It can be inferred from Eq. \ref{eq5} that the resulting feature $y$ at each position is a weighted sum of the features across all positions and original features. The similar semantic features achieve mutual benefits, improving intra-class compact and semantic consistency, which presents highlighting region in the similarity map. Code are available at \url{https://github.com/binyisu/cosine-nonlocal}. 

\subsection{Defect detection}
In this paper, we use Faster RCNN+FPN \cite{2017FPN} as the base detector to detect multi-scale defects in EL near-infrared images such as crack, finger interruption, and black core, while integrating a novel Bidirectional Attention Feature Pyramid Network (BAFPN) into RPN. We firstly introduce the proposed BAFPN block, then illustrate the details of proposed defect detection architecture BAF-Detector.

\subsubsection{Bidirectional Attention Feature Pyramid Network (BAFPN)}
To obtain more refined fusion features of pyramidal layers, a novel Bidirectional Attention Feature Pyramid Network (BAFPN) is designed by combining the novel multi-head cosine non-local attention module with top-down and bottom-up feature pyramid networks through bidirectional cross-scale connections. BAFPN can accomplish informative feature aggregation of pyramidal layers, and improve the robustness of the network to scales. In BAFPN, multi-head cosine non-local attention module plays a key role in feature refinement.

As shown in Fig. \ref{fig6}, multi-scale features are aggregated by the BAFPN block at different resolutions. Formally, given a list of multi-scale features (C2, C3, C4, C5), the goal of BAFPN block is to find a transform function that can effectively aggregate different features and output a list of new refined features. The top-down and bottom-up feature fusion process is defined as:

\begin{figure*}[!t]
	\centering
	\includegraphics[width=16cm]{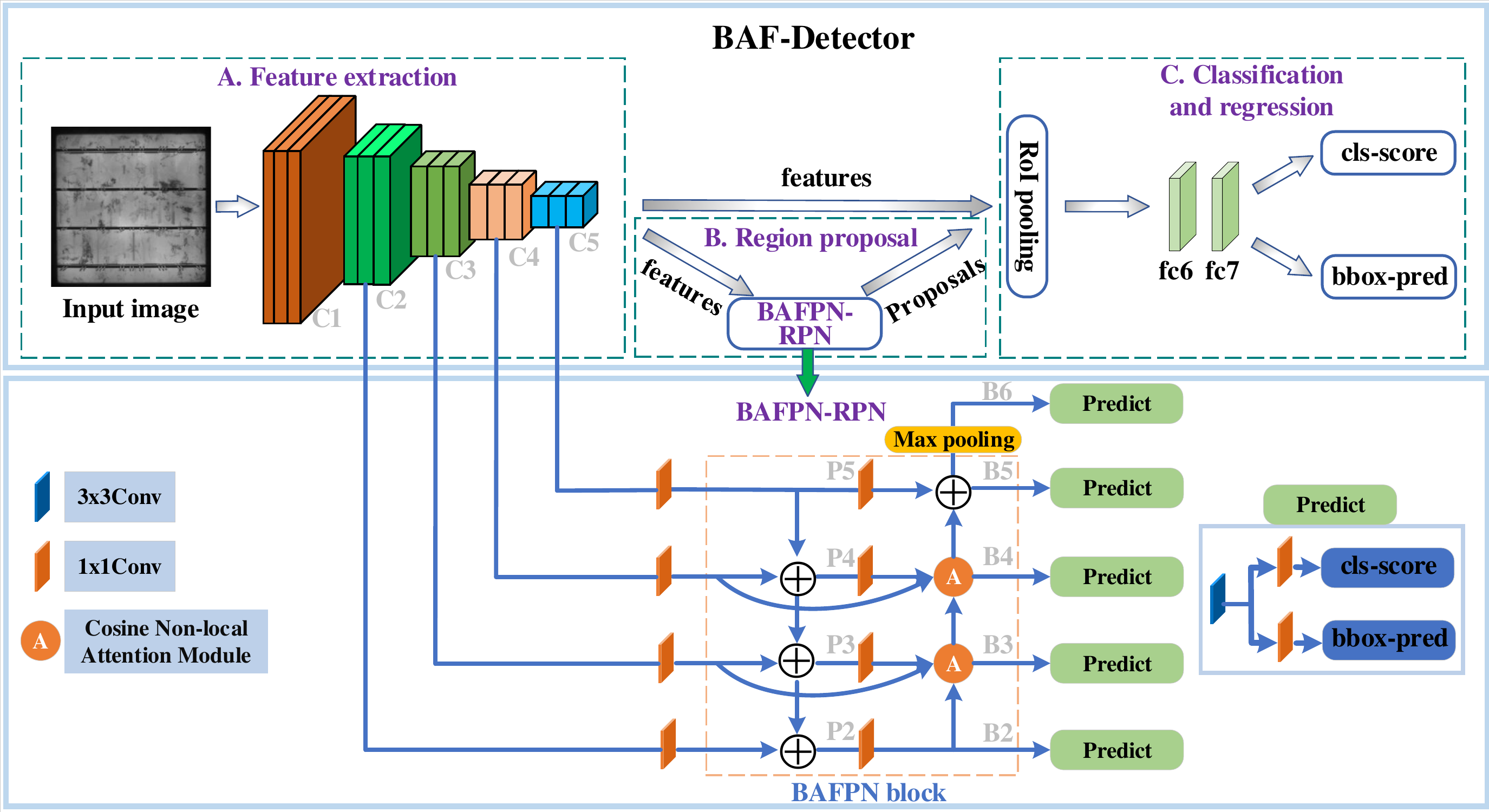}
	\caption{The architecture of the proposed Bidirectional Attention Feature Pyramid Network Detector (BAF-Detector).}\label{fig6}
\end{figure*}

\begin{equation}\label{eq6}
	\begin{split}
		P5=conv(C5),P4=up(P5)+conv(C4),\;\;\;\;\;\;\\P3=up(P4)+conv(C3),P2=up(P3)+conv(C2)
	\end{split}
\end{equation}
\begin{equation}\label{eq7}
	\begin{split}
		B2=conv(P2),\;\;\;\;\;\;\;\;\;\;\;\;\;\;\;\;\;\;\;\;\;\;\;\;\;\;\\B3=A(conv(C3),conv(P3),down(B2)),\;\;\;\;\;\;\\B4=A(conv(C4),conv(P4),down(B3)),\;\;\;\;\;\;\\B5=conv(P5)+down(B4),B6=max\;pool(B5)\;
	\end{split}
\end{equation}
where $up$ represents upsampling, $down$ represents downsampling, $max\;pool$ represents max pooling and $conv$ is a convolutional operation for feature filtering.  $A(\cdot)$ is the operation of proposed multi-head cosine non-local attention module. Rich textural features in lower layers by bottom-up path augmentation shorten the information path between lower layers and topmost feature \cite{2018PANet}. Moreover, the outputs of two middle layers ($B3$ and $B4$) are refined by attention mechanism, which will highlight the defect feature and suppress the complex background feature during the PV cell EL image defect detection task. The reason why attention modules are only used in the middle two layers of BAFPN is that on the one hand the multi-head cosine non-local attention module has a heavy computational burden. If it is used in every layer of FPN, it will cause our sever out of memory. On the other hand, the middle two layers of FPN are a compromise between useful semantic features and texture features, then using the attention module to refine the features of these two layers will maximize the benefits. 

By above computation, the outputs are the features after fusion of different scales. The pyramidal features fusion can increase the robustness to scale variance, all layers of our pyramid share similar semantic features.

\subsubsection{Defect Detection Architecture}
As shown in Fig. \ref{fig6}, the architecture of the proposed BAF-Detector is divided into three parts: feature extraction, region proposal (BAFPN-RPN), classification and detection. Firstly, ImageNet-pretrained ResNet101 \cite{2016Resnet} is employed as the backbone network to extract the features of the EL image. Secondly, the multi-level features will be passed to the proposed BAFPN-RPN module, which will generate many region proposals that may contain a defect. 
Thirdly, each proposal generated at different scale levels is mapped to the corresponding level features of ResNet101, next the Region of Interest (RoI) will resize the mapped features to a fixed-size vector. Finally, two fully connected layers is appended before two output layers: cls-score layer outputs classification scores of $K$ object classes attach a ``background" class, bbox-pred layer outputs the predicted positions of the bounding-box for the corresponding $K$ object classes. In BAF-Detector, each predicted head of BAFPN-RPN can be optimized to regression by the following loss function:
\vspace{-1em}

\begin{equation}\label{eq:7}
	\begin{split}
		L(\{p_i\},\{t_i\})=\frac1{N_{cls}}\sum_iL_{cls}(p_i,p_i^\ast)+\\
		\lambda\frac1{N_{reg}}\sum_ip_i^\ast L_{reg}(t_i,t_i^\ast)
	\end{split}
\end{equation}
where $i$ is the index of an anchor box \cite{2017Faster} in a mini-batch and $p_i$ is the predicted probability of anchor box $i$ containing a defect. If the anchor box is predicted to contain a defect, the ground-truth label $p_i^\ast$ is 1, otherwise $p_i^\ast$ is 0. $t_i$ is a vector representing the 4 parameterized coordinates of the predicted bounding box, and $t_i$ is that of the ground-truth box associated with a defective anchor box. The term $p_i^\ast L_{reg}$ means the regression loss is activated only for defective anchor $(p_i=1)$ and is disabled otherwise $(p_i=0)$. The outputs of the cls\_score layer and bbox\_pred layer consist of $p_{i}$ and $t_{i}$ respectively. The classification loss $L_{cls}$  is log loss over two classes (defect or not defect). For the regression loss, we use $ L_{reg}(t_i,t_i^\ast)=R(t_i-t_i^\ast)$ where $R$ is the robust loss function (smooth L1) defined as follows:

\begin{equation}\label{eq:8}
	smooth_{L1}(x)=\left\{\begin{array}{l}0.5x^2\;\;\;\;\;\;\;\;\;\;if\;\vert x\vert<1\\\vert x\vert-0.5\;\;\;\;\;otherwise\end{array}\right.
\end{equation}

The term $p_i^\ast L_{reg}$ means the regression loss is activated only for defective anchor boxes $(p_i^\ast=1)$ and is disabled otherwise $(p_i^\ast=0)$. Referring to Faster RCNN+FPN, the two terms are normalized by $N_{cls}=256$ and $N_{reg}=2400$, and weighted by a balancing parameter $\lambda=10$, thus both $cls$ and $reg$ terms are roughly equally weighted.

For bounding box regression, the parameterizations of the 4 coordinates is defined as follows:
\begin{equation}\label{eq:9}
	t_x=\frac{x-x_a}{w_a},\;\;\;t_y=\frac{y-y_a}{h_a}
	\vspace{-1em}
\end{equation}

\begin{equation}\label{eq:10}
	t_w=\log(\frac w{w_a}),\;\;\;t_h=\log(\frac h{h_a})
	\vspace{-1em}
\end{equation}

\begin{equation}\label{eq:11}
	t_x^\ast=\frac{x^\ast-x_a}{w_a},\;\;\;t_y^\ast=\frac{y^\ast-y_a}{h_a}
	\vspace{-1em}
\end{equation}

\begin{equation}\label{eq:12}
	\vspace{-1em}
	t_w^\ast=\log(\frac{w^\ast}{w_a}),\;\;\;t_h^\ast=\log(\frac{h^\ast}{h_a})
	\vspace{1em}
\end{equation}
where $x$, $y$, $w$ and $h$ donate the center coordinates, width and height of the box. Variables $x$, $x_a$, and $x^*$ are for the predicted box, anchor box and ground-truth box, respectively ($y$, $w$, $h$ are same as this). Otherwise, this can be thought of as bounding-box regression from an anchor box to a nearby ground-truth box. 

\subsubsection{Momentum-based Parameter Optimization}
The purpose of updating weight parameters through back propagation is to minimize the loss. When training the BAF-Detector, the loss is viewed as a function of the weight parameters, CNN need to calculate the partial derivative of the loss corresponding to each weight parameter, and then use the momentum-based gradient descent method to iteratively update weights in the direction of the fastest gradient descent, until the conditions for the parameter to stop updating are satisfied. For the momentum optimization method, when the parameter is updated, the previous update direction is retained to a certain extent, and the final update direction is fine-tuned using the gradient of the current batch. The current gradient is accelerated by accumulating the previous momentum. In short, if the gradient direction is unchanged, the parameter update will become faster. Otherwise, the update parameter will become slower. Thus, it can speed up the convergence and reduce the parameter oscillation.

%In mini-batch gradient descent, if the number of samples selected for training is relatively small, the loss will decrease in an oscillating manner. For the momentum optimization method, if the gradient direction is unchanged, the parameter update will become faster. When the gradient changes, the update parameter will become slower. Thus, it can speed up the convergence and reduce the shock. When the parameter is updated, the previous update direction is retained to a certain extent, and the final update direction is fine-tuned using the gradient of the current batch. In short, the current gradient is accelerated by accumulating the previous momentum.

\section{EXPERIMENTAL RESULTS}
In this section, experimental steps such as dataset construction, evaluation matrix, implementation details, quantitative experimental evaluation, time efficiency evaluation, and ablation studies are presented.  To understand and show the
improvement directly, the visualization of similarity map is used as a
powerful tool to support experimental results. 

\subsection{Dataset}
We evaluate the experimental results of our proposed BAF-Detector on our PV cell EL image dataset. In this dataset, 2129 EL defective images and 1500 defect-free images with raw resolution of 1024$\times$1024 are used to evaluate the classification and detection effectiveness of our proposed BAF-Detector. Table \ref{table_1} shows the raw dataset distribution of training data and testing data. Firstly, the dataset is divided into two types: defective images and defect-free images, which are used to evaluate the classification performance of the proposed model. Secondly, the dataset is divided in details according to the categories of the defect. The reason why we classify defects within three main categories, namely crack, finger interruption, and black core defect is that these three kinds of defects occur frequently, and we can build a balanced dataset distribution, which is the key point that the deep learning model can be trained to achieve good performance. We use electroluminescence (EL) technology \cite{2009Fuyuki} to visualize the internal defects of PV cells. 2129 defective images are selected from 150,000 samples of PV cell images. Among them, crack, finger interruption, and black core defects are the most frequent ones. Other types of defect occur rarely, which will cause an unbalanced distribution of the dataset. Therefore, this article mainly classifies these three types of defects.

For the ground truth of different defects, a dataset annotation tool (LabelImg) is used to label the EL image dataset. It just needs a rectangular box to tightly surround the defect and do not need too much expert experience. The rectangular box will reflect the specific location and class of the defect. Moreover, annotations are saved as XML files in PASCAL VOC format. The standard PASCAL VOC format can ensure the fairness comparison between different detectors. It is very important for the validation of the performance of our proposed model.

\subsection{Evaluation metrics}
The classification of our model is evaluated by the Precision (P), Recall (R) and F-measure (F). Moreover, Average Precision (AP), mean Average Precision (mAP) and Mean Intersection over Union (MIoU) are applied to assess defect detection results. Parameters number and frames per second (FPS) are the metrics used to asses the time efficiency. The aforesaid metrics are defined as:

\begin{equation}\label{eq:13}
	Precision=\frac{TP}{TP+FP}\times 100\%
\end{equation}

\begin{equation}\label{eq:14}
	Recall=\frac{TP}{TP+FN}\times 100\%
\end{equation}

\begin{equation}\label{eq:15}
	F-measure=\frac{2\times Precision\times Recall}{Precision+Recall}
\end{equation}
\begin{equation}\label{eq:16}
	IoU=\frac{DetectionResult\cap GroundTruth}{DetectionResult\cup GroundTruth}
\end{equation}
where $TP$ and $FN$ are the number of defect images, which are predicted to be correct or not; $FP$ represents the number of non-defect, which is misclassified; $DetectionResult$ is the defect detected box of the detector; $GroundTruth$ is the defect annotation box.

\subsection{Implementation Details}
The experiments are conducted on a work station with a Intel Core i7-10700 CPU and a NVIDIA GeForce RTX 2070 SUPER. The pre-trained model on Imagenet is used to initialize the ResNet101, which can accelerate convergence of the network. The learning-rate is set to 0.001. The class number is set to 3. Due to the limitation of the GPU memory, the batch size is set to 1. The max iteration is fixed to 40000, which can ensure the fully loop through of the PV cell EL training data. In BAFPN-RPN, the stride of $max\;pooling$ is 2. Moreover, a scale level is assigned with an anchor, which has three aspect ratios [0.5, 1, 2] at each level. The anchors is defined to have areas of $\{$$32^2$,$64^2$,$128^2$,$256^2$,$512^2$$\}$ pixels on $\{$B2,B3,B4,B5,B6$\}$ respectively \cite{2017FPN}, and anchors with different scales are responsible for predicting defects of different sizes. In the experiments, all input EL images are extended to three channels and resized to 600$\times$600 pixels. The detailed information and hyperparameters of the BAF-Detector are presented in Table \ref{table_2}, which can help the readers to better understand our method.

Moreover, in multi-head cosine non-local attention module, factors $\alpha$, $\beta$ and $\lambda$ are employed to balance the importance of the three inputs. For example, as shown in Fig. \ref{fig6}, the inputs of the second multi-head cosine non-local attention module ($B3$) are $C3$, $P3$ and $B2$. The simple method to determine $\alpha$, $\beta$ and $\lambda$ is $\alpha=\beta=\lambda$, which shows that the three inputs are equally important \cite{Bhojanapalli2020,RCAG2021}. As for $\alpha=\beta=\lambda =0.5$, we refer to the fact that when performing multi-scale feature summation in FPN \cite{2017FPN}, the balance value of each input is 0.5, thus we also select 0.5.

\begin{table}[]
	\renewcommand{\arraystretch}{1.1}
	\caption{Dataset Distribution. Ck, Fr and Bc: the Number of Crack Defect, Finger Interruption Defect and Black Core Defect respectively.}
	\centering
	\label{table_1}
	\scriptsize
	\resizebox{\columnwidth}{!}{
	\begin{tabular}{lccccc}
		\hline\hline
		{\color[HTML]{31353B} }                                                                                & {\color[HTML]{31353B} }                                                                            & {\color[HTML]{31353B} }                                                                              & \multicolumn{3}{c}{{\color[HTML]{31353B} Number of defects}}                           \\ \cline{4-6} 
		\multirow{-2}{*}{{\color[HTML]{31353B} \begin{tabular}[c]{@{}l@{}}Dataset \\ (EL image)\end{tabular}}} & \multirow{-2}{*}{{\color[HTML]{31353B} \begin{tabular}[c]{@{}c@{}}Defective\\ image\end{tabular}}} & \multirow{-2}{*}{{\color[HTML]{31353B} \begin{tabular}[c]{@{}c@{}}Defect-free\\ image\end{tabular}}} & {\color[HTML]{31353B} Ck}   & {\color[HTML]{31353B} Fr}   & {\color[HTML]{31353B} Bc}  \\ \hline
		{\color[HTML]{31353B} Training}                                                                        & {\color[HTML]{31353B} 847}                                                                         & {\color[HTML]{31353B} /}                                                                             & {\color[HTML]{31353B} 452}  & {\color[HTML]{31353B} 592}  & {\color[HTML]{31353B} 251} \\
		{\color[HTML]{31353B} Testing}                                                                         & {\color[HTML]{31353B} 1282}                                                                        & {\color[HTML]{31353B} 1500}                                                                          & {\color[HTML]{31353B} 685}  & {\color[HTML]{31353B} 1249} & {\color[HTML]{31353B} 272} \\
		{\color[HTML]{31353B} Total}                                                                           & {\color[HTML]{31353B} 2129}                                                                        & {\color[HTML]{31353B} 1500}                                                                          & {\color[HTML]{31353B} 1137} & {\color[HTML]{31353B} 1841} & {\color[HTML]{31353B} 523} \\ \hline\hline
	\end{tabular}}
\end{table}

% Please add the following required packages to your document preamble:
% \usepackage{multirow}
\begin{table}[]
	\renewcommand{\arraystretch}{1.3}
	\caption{Hyperparameters During BAF-Detector Training and Testing}
	\centering
	\label{table_2}
	%	\scriptsize
	\resizebox{\columnwidth}{!}{
		\begin{tabular}{ccccc}
			\hline\hline
			\multirow{4}{*}{Training} & DECAY\_STEP                    & DECAY\_FACTOR             & WEIGHT\_DECAY                 & MOMENTUM       \\
			& 15000, 30000                   & 10                  & 0.0001                & 0.9                \\
			& IMG\_SHORT\_SIDE\_LEN      & IMG\_MAX\_LENGTH     & BASE\_ANCHOR\_SIZE\_LIST       & BATCH\_SIZE          \\
			& 600                        & 1000                 & {[}32, 64, 128, 256, 512{]}         & 1                    \\ \hline
			\multirow{2}{*}{Testing}  & \multicolumn{2}{c}{SHOW\_SCORE\_THRSHOLD} & \multicolumn{2}{c}{NMS\_IOU\_THRESHOLD} \\
			& \multicolumn{2}{c}{0.3}                           & \multicolumn{2}{c}{0.5}                               \\ \hline\hline
	\end{tabular}}
\end{table}

\subsection{Evaluation}
The proposed BAF-Detector is compared with YOLOv3, Faster RCNN, and Faster RPAN-CNN in the classification and detection performance on the PV cell EL image dataset, which are presents in Table \ref{table_3}. 

\subsubsection{Classification Evaluation}
In terms of the defective images classification results, the proposed BAF-Detector achieves a better performance than other detectors on each quantitative evaluation metric, such as precision (99.21$\%$), recall (98.20$\%$) and F-measure (98.70$\%$). The high recall rate of the proposed BAF-Detector ensures that the defective PV cell is not easy to be missed during the intelligent manufacturing process, which is essential to the high-quality production of PV cells. 

\begin{figure}[!t]
	\centering
	\includegraphics[width=8.4cm]{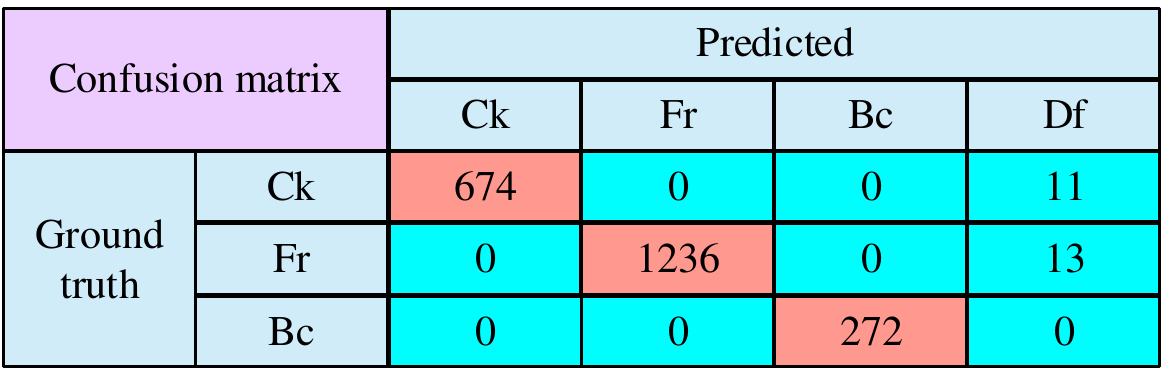}
	\caption{Confusion matrix for the classification evaluation. Df represents defect-free.}\label{fig6-1}
\end{figure}

As shown in Fig. \ref{fig6-1}, the confusion matrix is reported to evaluate the classification performance of the PV cell. The confusion matrix is given for the three types of defects: crack, finger interruption, and black core. In confusion matrix, 11 crack defects are misclassified as defect-free, which accounts for 1.61$\%$ of the total number of the crack defects. 13 finger interruption defects are misclassified as defect-free, which accounts for 1.04$\%$ of the total number of finger interruption defects. Moreover, all black core defects are classified correctly. The proposed CNN model makes less prediction error, a small proportion of defects are misclassified as defect-free. It is not hard to see that BAF-Detector achieves a good classification performance of PV cell defects.

\subsubsection{Detection Evaluation}
In terms of the proposed model in defect detection performance, mAP and MIoU are employed to evaluate the quantitative comparisons. Table \ref{table_3} shows the detection results of regression-based detector (YOLOv3) and region-based detectors (the others) with same training and testing EL image dataset. With embedding the proposed BAFPN block into RPN in Faster RCNN+FPN, the BAF-Detector outperforms other detectors in the defect detection effects. 

%The P/R curves of different defects using our proposed BAF-Detector are shown in Fig. \ref{fig7}. The detection result of the black core defect is the best, the AP value reaches 100$\%$. Black core defect has a single blob-like shape and a large scale, which causes it easy to be detected. The detection result of finger interruption defect is also easier to be detected, due to the vertical strip distribution. The detection result of crack defect is the worst, because of the random shapes and different scales especially for micro-crack, which bring some challenges to detect it in \textcolor{red}{PV} cell EL images. 

\begin{figure}[!t]
	\centering
	\includegraphics[width=8.4cm]{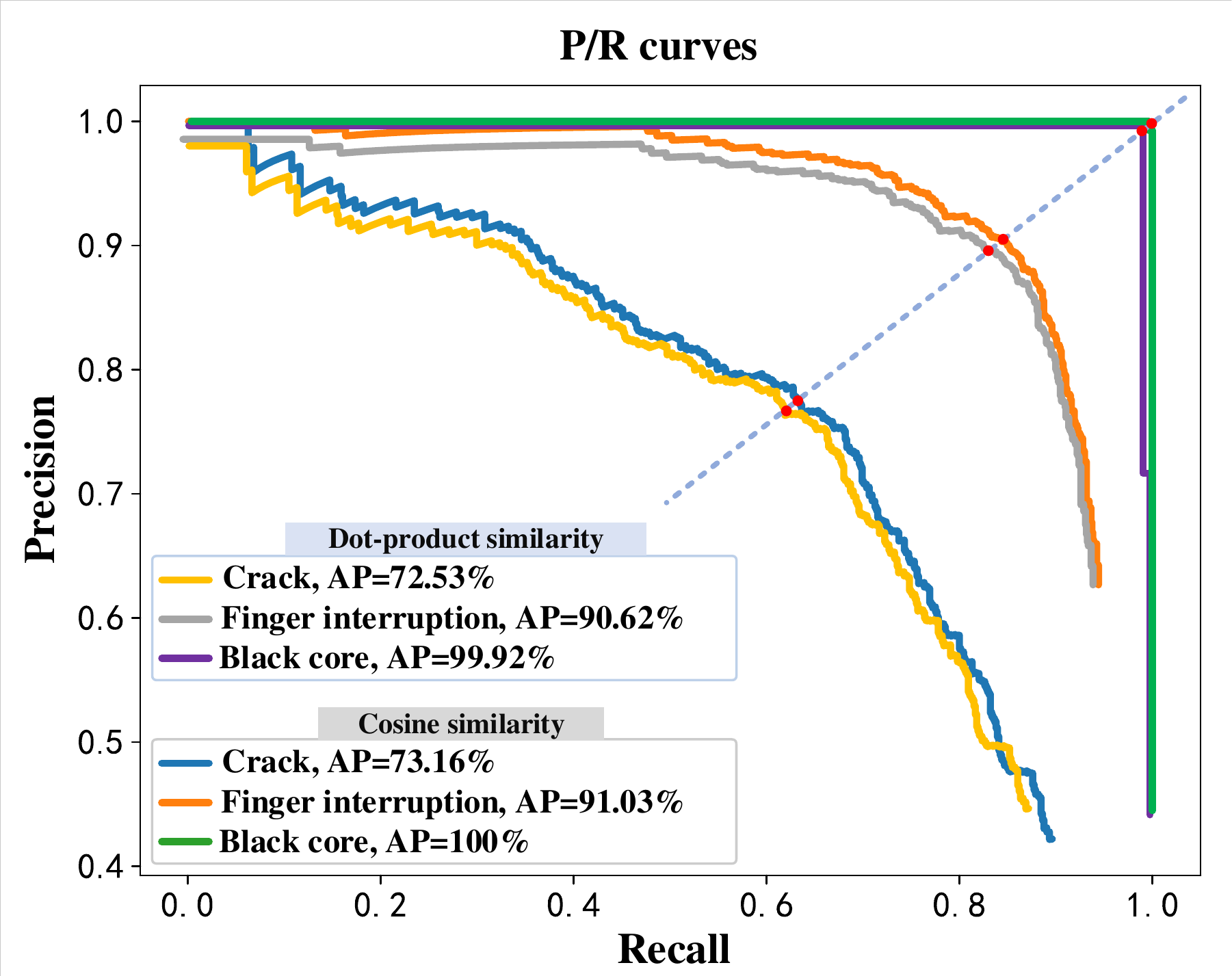}
	\caption{The P/R curves of different similarity calculation methods for three types of defects.}\label{fig7}
\end{figure}

The P/R curves of dot-product similarity and cosine similarity used in our proposed BAF-Detector are shown in Fig. \ref{fig7}. The area enclosed by the curve, the horizontal axis (recall) and the vertical axis (precision) is equal to the value of the average precision (AP) for each type of defect. The dotted line represents $precision=recall$. The intersection of it and the P/R curve represents the balance point (red dot). The larger the value of the balance point, the better the effect of the corresponding detection method. As can be seen from the Fig. \ref{fig7}, comparing with dot-product similarity and cosine similarity used in BAF-detector, cosine similarity achieves better quantitative results. 
	
Moreover, as can be seen from Fig. \ref{fig7}, the AP value of the black core defect is close to 100$\%$. This is due to the large scale and simple texture of the black core defect, which presents as a black cluster area. The detection result of finger interruption defect is relatively good, due to the sharp contrast and the fixed shape. The crack defect is hard to be detected, because of various scales, different shapes and the complex background disturbance.

\begin{figure}[!t]
	\centering
	\includegraphics[width=8.4cm]{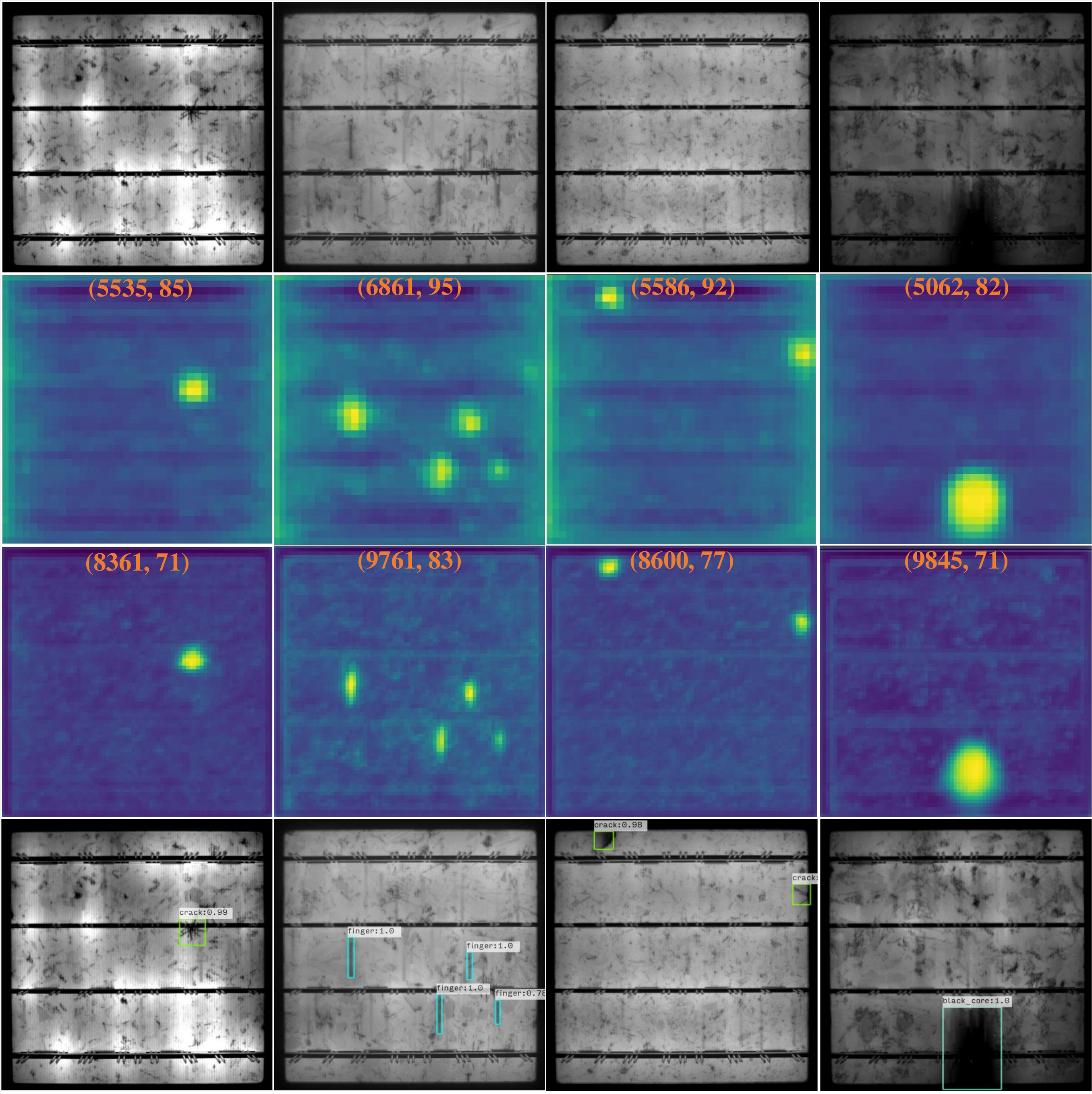}
	\caption{Visualization of intermediate and final detection results. The first row is the input, the second row is the similarity map of B4 layer, the third row is the similarity map of B3 layer, and the fourth row is the output. The orange value represents the contrast and brightness of each similarity map.}\label{fig8}
\end{figure}

The similarity map of the multi-head cosine non-local attention module can be viewed as a feature visualization tool, which can help explain the effectiveness of the BAFPN. As shown in Fig. 8, B3 and B4 belong to BAFPN, if the cosine similarity maps of B3 and B4 include the small defect feature, it illustrates that BAFPN is effective to boost the small defect feature transfer, and prevent the feature vanishment as the network deepens. Fig. \ref{fig8} shows some intermediate and final detection results. As can be seen, the similarity maps of low layer B3 or deep layer B4 in BAFPN retain the informative features of multi-scale defects during the network deepens. At the same time, the noise background features are suppressed. Thus, we can conclude that the BAFPN boosts the bottom-up feature transfer as the network deepens. It enables the BAF-Detector to better detect multi-scale defects in EL images. Moreover, the contrast of the low-level similarity map in B3 layer is higher than that of the high-level similarity map in B4 layer, and the brightness is opposite, which shows that multi-head cosine non-local attention achieves better performance in higher resolution features that contains much more textural and spatial information.

% Please add the following required packages to your document preamble:
% \usepackage{multirow}
\begin{table}[]
	\renewcommand{\arraystretch}{1.3}
	\caption{The Experimental Results of Different Detectors on Our PV Cell EL Image Dataset.}
	\centering
	\label{table_3}
	\resizebox{\linewidth}{!}{
		\begin{tabular}{lccccccc}
			\hline\hline
			\multirow{2}{*}{Detectors} & \multicolumn{3}{c}{Classification} & \multicolumn{2}{c}{Detection} & \multirow{2}{*}{Parameters} & \multirow{2}{*}{FPS} \\ \cline{2-6}
			& P(\%)    & R(\%)   & F(\%)         & mAP(\%)       & MIoU(\%)      &                             &                      \\ \hline
			YOLOv3 \cite{2020Su}           & 86.73    & 97.30    & 91.71         & 78.79         & 64.96         & 58.73M                      & 11.36                \\
			Faster RCNN \cite{2020Su}      & 93.53    & 96.04   & 94.77         & 83.11         & 68.22         & 260.50M                     & 6.41                 \\
			Faster RPAN-CNN \cite{2020Su}  & 99.17    & 97.78   & 98.47   & 87.38         & 71.42         & 261.26M                     & 5.95                 \\
			Our BAF-Detector           & 99.21    & 98.20    & 98.70   & 88.07         & 73.29         & 120.87M                     & 7.75                 \\ \hline\hline
	\end{tabular}}
\end{table}

% Please add the following required packages to your document preamble:
% \usepackage{multirow}
% \usepackage[table,xcdraw]{xcolor}
% If you use beamer only pass "xcolor=table" option, i.e. \documentclass[xcolor=table]{beamer}
\begin{table}[tbp]
	\vspace{-1em}
	\renewcommand{\arraystretch}{1.3}
	\caption{The Ablation Studies of the Proposed BAF-Detector on Our PV Cell EL Image Dataset with Different Components. $\oplus$: Addition Operation is Used to Replace the Attention Module. mt: Middle Two Scale Levels, Attention Modules are Used in the Middle Two Levels of FPN. \xmark: Not Use Attention Module. \cmark: Use Attention Module.}
	\centering
	\label{table_4}
	\resizebox{\linewidth}{!}{
		\begin{threeparttable}	
			\begin{tabular}{llccccccc}
				\hline\hline
				&                             & \multicolumn{3}{c}{Classification} & \multicolumn{2}{c}{Detection} &                                 &                       \\ \cline{3-7}
				\multirow{-2}{*}{FPN} & \multirow{-2}{*}{Attention} & P(\%)     & R(\%)    & F(\%)       & mAP(\%)       & MIoU(\%)      & \multirow{-2}{*}{Parameters}    & \multirow{-2}{*}{FPS} \\ \hline
				top-down              & \xmark($\oplus$)                & 94.62     & 96.72    & 95.66    & 85.21         & 70.53         & 115.11M & 7.81                  \\
				top-down              & \cmark(dot-product,mt)       & 95.87       & 97.42      & 96.64         & 86.97           & 71.54           & 115.37M                             & 7.78                   \\
				top-down              & \cmark(cosine,mt)            & 96.18       & 97.86      & 97.01         & 87.14           & 72.56           & 115.37M                             & 7.79                   \\
				top-down+bottom-up    & \xmark($\oplus$)                & 95.13       & 97.07      & 96.09         & 85.67         & 72.32         & 119.87M                         & 7.75                  \\
				top-down+bottom-up    & \cmark(dot-product,mt)       & 98.58       & 97.91      & 98.24         & 87.69         & 72.58         & 120.87M                         & 7.69                  \\
				top-down+bottom-up    & \cmark(cosine,mt)            & 99.21     & 98.20     & 98.70    & 88.07         & 73.29         & 120.87M                         & 7.68                  \\ \hline\hline
			\end{tabular}
			
			%	\begin{tablenotes}
			%		\footnotesize
			%		\item[1] mt: middle two layers.
			%	\end{tablenotes}
	\end{threeparttable}}
\end{table}

\subsubsection{Overall Evaluation}
From above classification and detection evaluation of the proposed BAF-Detector, we can conclude that the proposed BAF-Detector outperforms other detectors in the EL image  defect classification and detection performance. The reason is that the fusion strategy (BAFPN) can employ the proposed multi-head cosine non-local attention module and the  top-down bottom-up FPN to highlight target feature, suppress complex background feature, and better guide pyramidal feature fusion, which is very beneficial for multi-scale defect classification and detection tasks under complex background interference in PV cell EL image dataset.

\subsubsection{Time-efficiency Evaluation}
The results of FPS is the average of 1282 testing defective EL image, which is conducted on a work station with a GPU. As we can see from Table \ref{table_3}, the FPS of a detector is inversely proportional to the parameter number. Our BAF-Detector achieves 7.75 frame per second (FPS). Although our method is not the fastest, but it outperforms other detectors in defect inspection accuracy that is essential to the industrial production.

\vspace{-1em}
\subsection{Ablation Studies}

\subsubsection{Effect of Attention Module}
To confirm the effectiveness of the proposed multi-head cosine non-local attention module, the strategy of add or replace the attention module is employed for ablation studies. As illustrated in Table \ref{table_4}, by comparing two similarity calculation methods of non-local (dot-product) and cosine non-local (cosine), it can be demonstrated that cosine similarity outperforms dot-product similarity in the PV cell defect inspection task. The visualization of similarity maps are shown in Fig. \ref{fig5}. For the second row of the comparisons, defect regions can be highlighted more clearly by cosine similarity map, and the background is more fully suppressed. Moreover, the performance of adding attention module is better than removing it.

\subsubsection{Effect of Bidirectional Attention Feature Pyramid Network (BAFPN)}
Ablation studies are employed to verify the effectiveness of the proposed BAFPN. The ablation experiments are evaluated by changing the structure of the FPN (top-down, top-down+bottom-up). The effect of the top-down bottom-up FPN with attention module (BAFPN) can be seen in the last two rows in Table \ref{table_4}. By comparing the proposed BAFPN with other structures, we find that BAFPN achieves a better performance than the experimental results of the first four rows, which illustrates that the BAFPN block outperforms others in terms of multi-scale defect detection. Moreover, as illustrated in the first three rows and the last three rows, FPN with top-down+bottom-up performs better than FPN with top-down under the condition of consistent attention modules. Take mAP column as an example, top-down+bottom-up architecture achieves 0.46$\%$, 0.72$\%$ and 0.93$\%$ improvement comparing with top-down architecture respectively. It verifies that bidirectional multi-scale feature fusion is better than a single direction. However, BAFPN has 4.77$\%$ more parameters than original FPN, which slightly increases computational burdens.

\section{CONCLUSION}
In this paper, a novel Multi-head Cosine Non-local Attention module is proposed to highlight the defect feature and suppress the complex background feature. The novel attention module is also used to construct an efficient multi-scale feature fusion block BAFPN, which combines with RPN in Faster RCNN+FPN to form a defect detection model BAF-Detector. The multi-scale defects in PV cell EL image are effectively detected under the disturbance of the complex background. The proposed BAF-Detector outperforms other detectors in each quantitative metric, which demonstrates that 
our proposed deep learning model has certain advantages in defect inspection, and provides a practical solution for research and applications of PV cell defect detection.

There are also some limitations for our BAF-Detector. For example, the feature balance factors $\alpha$, $\beta$ and $\gamma$ in attention module are set mutually, which takes time and efforts. In the future, adaptive methods will be researched to automatically determine these factors.

\section*{Appendix}

Here are some details for the intelligent defect detection system. 1) The intelligent defect detection system needs to be grounded to prevent harm caused by current leakage. And the leakage circuit breaker is also employed to ensure the safe and efficient operation of the system. 2) It is necessary to ensure that the image acquisition conditions do not change, which is necessary to the normal and long-term operation of the proposed algorithm. Please refer to Reference \cite{2019Su} for the image acquisition conditions.

% References

\bibliographystyle{Bibliography/IEEEtranTIE}
\normalem
%\bibliography{Bibliography/IEEEabrv,Bibliography/BIB_xx-TIE-xxxx}\ %IEEEabrv instead of IEEEfull

\vspace{-1cm}
\begin{IEEEbiography}[{\includegraphics[width=1in,height=1.25in,clip,keepaspectratio]{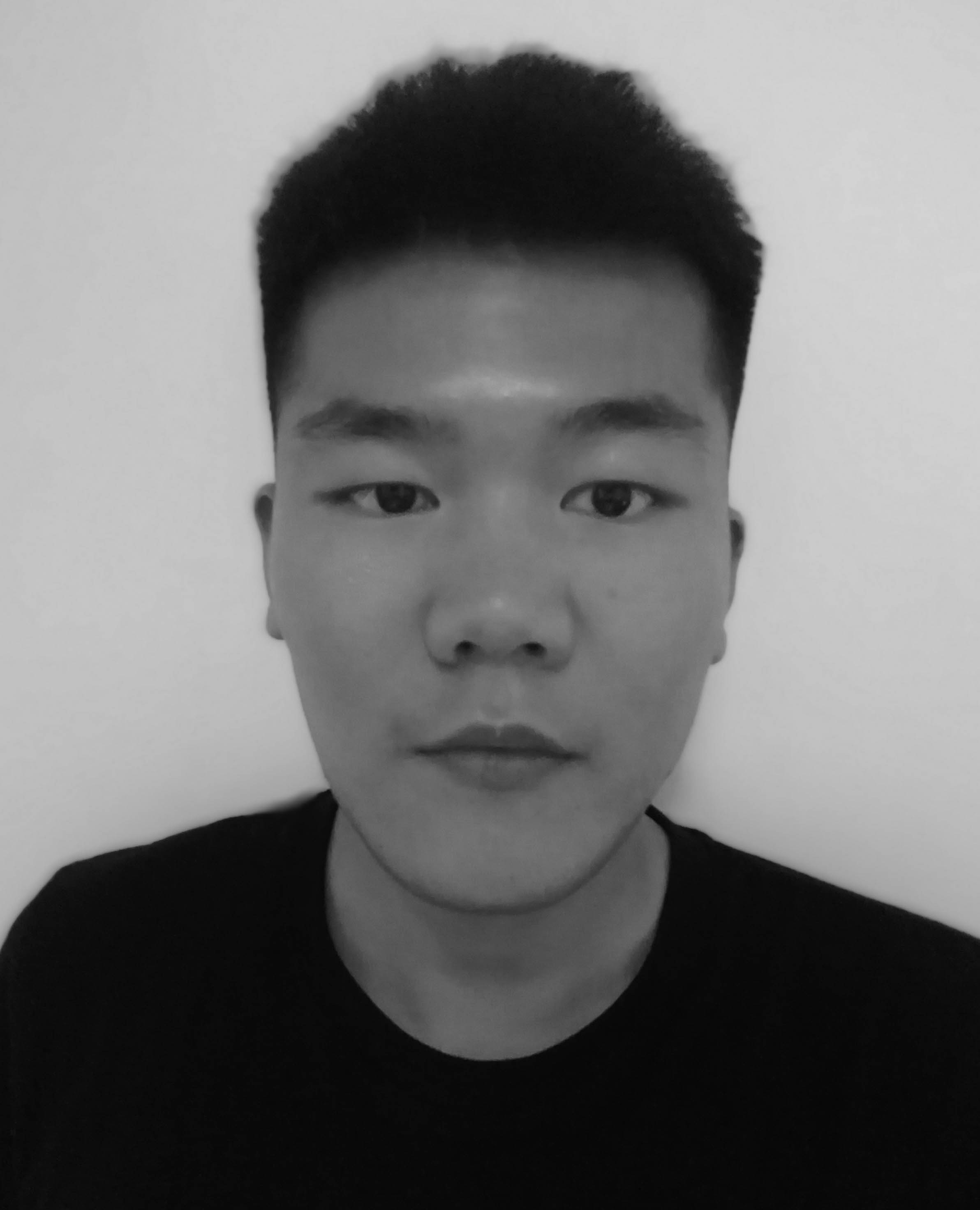}}]
	{Binyi Su} received the B.S. degree in intelligent 
	science and technology from the Hebei
	University of Technology, Tianjin, China, in 2017, and the M.S degree in control engineering from the Hebei
	University of Technology, Tianjin, China, in 2020.
	
	He is currently pursuing the Ph.D. degree in computer science and technology from Beihang University, Beijing, China.
	His current research interests include computer vision and pattern recognition, machine learning and artificial intelligence, industrial image defect detection.
\end{IEEEbiography}

\begin{IEEEbiography}[{\includegraphics[width=1in,height=1.25in,clip,keepaspectratio]{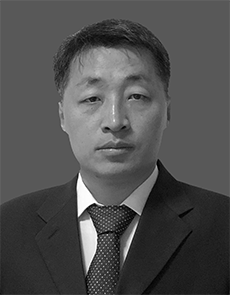}}]{Haiyong Chen}
	received the M.S. degree in detection technology and automation from the Harbin University of Science and Technology, Harbin, China, in 2005, and the Ph.D. degree in control science and engineering from the Institute of Automation, Chinese Academy of Sciences, Beijing, China, in 2008. 
	
	He is currently a Professor with the School of Artificial Intelligence and Data Science, Hebei University of Technology, Tianjin. He is also an expert in the field of photovoltaic cell image processing and automated production equipment. His current research interests include image processing, robot vision, and pattern recognition.
\end{IEEEbiography}

\begin{IEEEbiography}[{\includegraphics[width=1in,height=1.25in,clip,keepaspectratio]{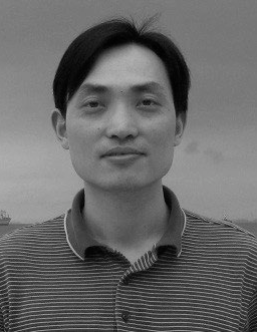}}]{Zhong Zhou} (Member, IEEE) received the B.S. degree in material physics from Nanjing University in 1999 and the Ph.D. degree in computer science and technology from Beihang University, Beijing, China, in 2005. 
	
He is currently a Professor and the Ph.D. Adviser with the State Key Laboratory of Virtual Reality Technology and Systems, Beihang University. His main research interests include virtual reality, augmented reality, mixed reality, computer vision, and artificial intelligence. He is a member of ACM and CCF.
\end{IEEEbiography}

\end{document}